\documentclass[english]{llncs}
\usepackage[latin9]{inputenc}
\usepackage{float}
\usepackage{textcomp}
\usepackage{amsmath}
\usepackage{amssymb}
\usepackage{stackrel}
\usepackage{graphicx}
\usepackage{setspace}

\makeatletter

\pdfpageheight\paperheight
\pdfpagewidth\paperwidth

\floatstyle{ruled}
\newfloat{algorithm}{tbp}{loa}
\providecommand{\algorithmname}{Algorithm}
\floatname{algorithm}{\protect\algorithmname}

\usepackage{numcompress}

\@ifundefined{showcaptionsetup}{}{%
 \PassOptionsToPackage{caption=false}{subfig}}
\usepackage{subfig}
\makeatother

\usepackage{babel}
\begin{document}

\title{\noindent Sparse Spectrum Gaussian Process for Bayesian Optimization }

\author{Ang Yang\and Cheng Li\and Santu Rana\and Sunil Gupta\and Svetha
Venkatesh}

\institute{Applied Artificial Intelligence Institute ($\mathrm{A^{2}I^{2}}$),\\
Deakin University, Geelong, Australia\\
\{leon.yang, cheng.l, santu.rana, sunil.gupta,\\
 svetha.venkatesh\}@deakin.edu.com\\
}
\maketitle

\begin{abstract}
\noindent We propose a novel sparse spectrum approximation of Gaussian
process (GP) tailored for Bayesian optimization. Whilst the current
sparse spectrum methods provide desired approximations for regression
problems, it is observed that this particular form of sparse approximations
generates an overconfident GP, i.e. it produces less epistemic uncertainty
than the original GP. Since the balance between predictive mean and
the predictive variance is the key determinant to the success of Bayesian
optimization, the current sparse spectrum methods are less suitable
for it. We derive a new regularized marginal likelihood for finding
the optimal frequencies to fix this over-confidence issue, particularly
for Bayesian optimization. The regularizer trades off the accuracy
in the model fitting with targeted increase in the predictive variance
of the resultant GP. Specifically, we use the entropy of the global
maximum distribution from the posterior GP as the regularizer that
needs to be maximized. Since this distribution cannot be calculated
analytically, we first propose a Thompson sampling based approach
and then a more efficient sequential Monte Carlo based approach to
estimate it. Later, we also show that the Expected Improvement acquisition
function can be used as a proxy for the maximum distribution, thus
making the whole process further efficient. Experiments show considerable
improvement to Bayesian optimization convergence rate over the vanilla
sparse spectrum method and over a full GP when its covariance matrix
is ill-conditioned due to the presence of large number of observations.
\end{abstract}

\section{Introduction }

Bayesian optimization (BO) is a leading method for global optimization
for expensive black-box functions \cite{33,13,shahriari2015taking,A_Yang_PRICAI_2018}.
It is widely used in hyperparameter tuning of massive neural networks
\cite{14,springenberg2016bayesian}, some of which can take days to
train. It has also been used for optimization of physical products
and processes \cite{li2017rapid} where one experiment can cost days,
and experiments can also be expensive in terms of material cost. However,
there could be scenarios when a large number of observations is available
from priors or during the experiments. For example, in transfer learning,
where many algorithms \cite{pan2009survey,long2013transfer,joy2019flexible}
pool existing observations from source tasks together for use in the
optimization of a target task. Then even though the target function
is expensive, the number of observations can be large if the number
of source tasks is large and/or the number of observations from each
source is large. Another scenario where we may have a large number
of observations is when we deal with optimization of objective functions
which are not very costly. For example, consider the cases when Bayesian
optimization is performed using simulation software. They are often
used in the early stage of a product design process to reduce a massive
search space to a manageable one before real products are made. Whilst
evaluation, a few thousands may be feasible, but millions are not
because each evaluation can still take from several minutes to hours.
We term this problem as a semi-expensive optimization problem. Such
problems cannot be handled by the traditional global optimizers which
often require more than thousands of evaluations. Bayesian optimization
will also struggle, because its main ingredient, Gaussian process
(GP) does not scale well beyond few hundreds of observations. In
this paper, we address the scalability issue of GP for Bayesian optimization
for such scenarios where a large number of observations appear naturally. 

The scalability issue for Bayesian optimization has been previously
addressed in two main ways: 1) by replacing GP with a more scalable
Bayesian model, \emph{e.g.} using Bayesian neural network \cite{26}
or random forest \cite{hutter2011sequential}, or 2) by making sparse
approximation of the full GP. The latter is often desirable as it
still maintains the principled Bayesian formalism of GP. There are
many sparse models in the literature, such as fully independent training
conditional (FITC) \cite{snelson2006sparse,yang2018sparse} which
induces pseudo inputs to approximate the full GP, and variational
approximation \cite{titsias2009variational} which learns inducing
inputs and hyperparameters by minimizing the KL divergence between
the true posterior GP and an approximate one. Another line of work
involves approximating a stationary kernel function using a sparse
set of frequencies in its spectrum domain representation, e.g. sparse
spectrum Gaussian process (SSGP) \cite{quia2010sparse}. These methods
suffer from either variance under-estimation (i.e., over-confidence)
\cite{snelson2006sparse,quia2010sparse} or over-estimation \cite{titsias2009variational}
and thus may hamper BO as the balance between predictive mean and
variance is important to the success of BO. Recently, \cite{hensman2017variational}
has proposed variational Fourier features (VFF), which combines variational
approximation and spectral representation of GP together and plausibly
can approximate both mean and variance well. However, it is difficult
to extend VFF to multiple dimensional problems, since a) the number
of inducing variables grows exponentially with dimensions if the Kronecker
kernel is used, or b) the correlation between dimensions would be
ignored if an additive kernel is used. 

In this paper, we aim to develop a sparse GP model tailored for Bayesian
optimization. We propose a novel modification to the sparse spectrum
Gaussian process approach to make it more suitable for Bayesian optimization.
The main intuition that drives our solution is that while being over-confident
at some regions is not very critical to Bayesian optimization when
those regions have both low predictive value and low predictive variance.
However, being over-confident in the regions where either predictive
mean or predictive variance is high would be quite detrimental to
Bayesian optimization. Hence, a targeted fixing may be enough to make
the sparse models suitable for BO. An overall measure of goodness
of GP approximation for BO would be to look at the global maximum
distribution (GMD) \cite{39,40} from the posterior GP and check its
difference to that of the full GP. Fixing over-confidence in the important
regions may be enough to make the GMD of the sparse GP closer to that
of the full GP. The base method in our work (SSGP) is known to under-estimate
variance, which is why we need maximizing the entropy of GMD. Following
this idea, we add entropy of the maximum distribution as a new regularization
term that is to be maximized in conjunction with the marginal likelihood
so that the optimal sparse set of the frequencies are not only benefit
for model fitting, but also fixes the over-confidence issue from the
perspectives of the Bayesian optimization. 

We first provide a Thompson sampling approach to estimate the maximum
distribution for the sparse GP, and then propose a more efficient
sequential Monte Carlo based approach. This approach provides efficiency
as many Monte Carlo samples can be reused during the optimization
for the optimal frequencies. Moreover, the maximum distribution also
does not change much between two consecutive iterations of Bayesian
optimization as the GP differs by only one observation. Later, we
empirically show that Expected Improvement acquisition function can
be used as a proxy of the maximum distribution, significantly improving
the computational efficiency. We demonstrate our method on two synthetic
functions and two real world problems, one involving hyperparameter
optimization in a transfer learning setting and another involving
alloy design using a thermodynamic simulator. In all the experiments
our method provides superior convergence rate over standard sparse
spectrum methods. Additionally, our method also performs better than
the full GP when the covariance matrix faces ill-conditioning due
to large number of observations placed close to each other.

\section{Background}

\noindent We consider the maximization problem $\boldsymbol{x}^{*}=\text{argmax}_{\boldsymbol{x}\in\mathcal{X}}f(\boldsymbol{x})$,
where $f:\boldsymbol{x}\rightarrow\mathbb{R}$, $\mathcal{X}$ is
a compact subspace in $\mathbb{R}^{d}$, and $\boldsymbol{x}^{*}$
is the global maximizer. 

\subsection{Bayesian optimization}

Bayesian optimization includes two main components. It first uses
a probabilistic model, typically a GP, to model the latent function
and then constructs a surrogate utility function, or acquisition function
which encapsulates optimistic estimate of the goodness about the next
query. 

Gaussian process \cite{16} provides a distribution over the space
of functions and it can be specified by a mean function $\mu(\boldsymbol{x})$
and a covariance function $k(\boldsymbol{x},\boldsymbol{x}')$. A
sample from a GP is a function $f(\boldsymbol{x})\sim\mathcal{GP}(\mu(\boldsymbol{x}),k(\boldsymbol{x},\boldsymbol{x}'))$.
Without loss of generality, we often assume that the prior mean function
is zero function and thus GP can be fully defined by $k(\boldsymbol{x},\boldsymbol{x}')$.
The squared exponential kernel and the Mat$\acute{e}$rn kernel are
popular choices of $k$. 

In GP, the joint distribution for any finite set of random variables
are multivariate Gaussian distribution. Given a set of noisy observations
$\mathcal{D}_{t}=\{\boldsymbol{x}_{i},y_{i}\}_{i=1}^{t}$, where $y_{i}=f_{i}+\varepsilon_{i}$
with $\varepsilon_{i}\sim\mathcal{N}(0,\sigma_{n}^{2})$, the predictive
distribution of $y_{t+1}$ in GP follows a normal distribution $p(f_{t+1}|\mathcal{D}_{t},\boldsymbol{x}_{t+1})=\mathcal{N}(\mu(\boldsymbol{x}_{t+1}),\sigma^{2}(\boldsymbol{x}_{t+1}))$
with
\begin{align}
\mu(\boldsymbol{x}_{t+1}) & =\mathbf{k}^{T}\left[\mathbf{K}+\sigma_{n}^{2}\mathbf{I}\right]^{-1}\boldsymbol{Y}\label{eq:4}\\
\sigma^{2}(\boldsymbol{x}_{t+1}) & =k(\boldsymbol{x}_{t+1},\boldsymbol{x}_{t+1})-\mathbf{k}^{T}\left[\mathbf{K}+\sigma_{n}^{2}\mathbf{I}\right]^{-1}\mathbf{k}\text{,}
\end{align}
where $\mathbf{k}=\left[k(\boldsymbol{x}_{t+1},\boldsymbol{x}_{1}),k(\boldsymbol{x}_{t+1},\boldsymbol{x}_{2}),\cdots,k(\boldsymbol{x}_{t+1},\boldsymbol{x}_{t})\right]$,
$\boldsymbol{Y}=\{y_{i}\}_{i=1}^{t}$and $\mathbf{K}$ is the Gram
matrix. 

The posterior computation of GP involves the inversion of the Gram
matrix and it would be very costly for a large number of observations.
Sparse approximation is the usual way to reduce the computational
cost with slight reduction in modeling accuracy. In this paper we
focus on the SSGP for optimization purpose due to its simplification
and scalability, details of which is discussed in the following subsection. 

Once the GP has been built to model the latent function, we can construct
acquisition functions by combining the predictive mean and variance
of the posterior GP to find the next query. Some popular acquisition
functions include expected improvement (EI), and GP-UCB \cite{12,29}
. We use EI function since it can work well. The EI is defined as
the expected improvement over the current maximal value $f(\boldsymbol{x}^{+})$.
The analytic form of EI can be derived as \cite{33}
\[
\alpha_{EI}(\boldsymbol{x})=\begin{cases}
\sigma(\boldsymbol{x})Z\Phi(Z)+\sigma(\boldsymbol{x})\phi(Z) & \sigma(\boldsymbol{x})>0\\
0 & \sigma(\boldsymbol{x})=0
\end{cases},
\]

\noindent where $Z=\left[\mu(\boldsymbol{x})-f(\boldsymbol{x}^{+})\right]/\sigma(\boldsymbol{x}).$
$\Phi(Z)$ and $\phi(Z)$ are the cumulative distribution function
and probability density function of the standard normal distribution,
respectively. 

\subsection{Sparse spectrum Gaussian process}

Sparse Gaussian process often introduce inducing points to approximate
the posterior mean and variance of full GP whilst sparse spectrum
Gaussian process uses optimal spectrum frequencies to approximate
the kernel function. Briefly, according to the Bochner\textquoteright s
theorem \cite{bochner1959lectures}, any stationary covariance function
can be represented as the Fourier transform of some finite measure
$\sigma_{f}^{2}p(\mathbf{s})$ with $p(\mathbf{s})$ a probability
density as
\begin{equation}
k(\boldsymbol{x}_{i},\boldsymbol{x}_{j})=\int_{\mathbb{R}^{D}}e^{2\pi i\mathbf{\boldsymbol{s}}^{T}(\boldsymbol{x}_{i}-\boldsymbol{x}_{j})}\sigma_{f}^{2}p(\mathbf{s})d\mathbf{s},\label{eq:kernelspectrum}
\end{equation}

\noindent where the frequency vector $\mathbf{s}$ has the same length
$D$ as the input vector $\boldsymbol{x}$. In other words, a spectral
density entirely determines the properties of a stationary kernel.
Furthermore, Eq.(\ref{eq:kernelspectrum}) can be computed and approximated
as 
\begin{align}
k(\boldsymbol{x}_{i},\boldsymbol{x}_{j}) & =\sigma_{f}^{2}\mathbb{E}_{p(\mathbf{s})}\left[e^{2\pi i\mathbf{s}^{T}\boldsymbol{x}_{i}}(e^{2\pi i\mathbf{s}^{T}\boldsymbol{x}_{j}})^{*}\right]\nonumber \\
 & \simeq\frac{\sigma_{f}^{2}}{m}\stackrel[r=1]{m}{\mathbf{\boldsymbol{\sum}}}\mathrm{cos}\left[2\pi\boldsymbol{s}_{r}^{T}(\boldsymbol{x}_{i}-\boldsymbol{x}_{j})\right]\label{eq:kernelcos}\\
 & =\frac{\sigma_{f}^{2}}{m}\phi(\boldsymbol{x}_{i})^{T}\phi(\boldsymbol{x}_{j}).\label{eq:abc}
\end{align}
The Eq.(\ref{eq:kernelcos}) is obtained by Monte Carlo approximation
with symmetric sets $\{\mathbf{s}_{r},-\mathbf{s}_{r}\}_{r=1}^{m}$
sampled from $\mathbf{s}_{r}\sim p(\mathbf{s})$, where $m$ is the
number of spectral frequencies (features). To obtain the Eq.(\ref{eq:abc}),
we use the setting
\begin{multline}
\phi(\boldsymbol{x})=[\mathrm{cos}(2\pi\boldsymbol{s}_{1}^{T}\boldsymbol{x}),\,\mathrm{sin}(2\pi\boldsymbol{s}_{1}^{T}\boldsymbol{x}),\cdots,\\
\,\mathrm{cos}(2\pi\boldsymbol{s}_{m}^{T}\boldsymbol{x}),\,\mathrm{sin}(2\pi\boldsymbol{s}_{m}^{T}\boldsymbol{x})]^{T},
\end{multline}
which is a column vector of length 2$m$ containing the evaluation
of the $m$ pairs of trigonometric functions at $\boldsymbol{x}$.
Then, it is straightforward to compute the posterior mean and variance
of SSGP 
\begin{align}
\mu(\boldsymbol{x}_{t+1}) & =\phi(\boldsymbol{x}_{t+1})^{T}\mathbf{A}^{-1}\boldsymbol{\Phi}\boldsymbol{Y}\label{eq:8}\\
\sigma^{2}(\boldsymbol{x}_{t+1}) & =\sigma_{n}^{2}+\sigma_{n}^{2}\phi(\boldsymbol{x}_{t+1})^{T}\mathbf{A}^{-1}\phi(\boldsymbol{x}_{t+1}),\label{eq:9}
\end{align}

\noindent where $\boldsymbol{\Phi}=[\phi(\boldsymbol{x}_{1}),\ldots\,,\phi(\boldsymbol{x}_{t})]\in\mathbb{R}^{2m\times t}$
and $\mathbf{A}=\boldsymbol{\Phi}\boldsymbol{\Phi}^{T}+\frac{m\sigma_{n}^{2}}{\sigma_{f}^{2}}\mathbf{I}_{2m}$.
To select optimal frequencies, we can maximize the log marginal likelihood
$\mathrm{log}p(\boldsymbol{Y}|\varTheta)$, where $\varTheta$ is
the set of all hyperparameters in the kernel function and the spectrum
frequencies,
\begin{align}
\mathrm{log}p(\boldsymbol{Y}|\varTheta) & =-\frac{1}{2\sigma_{n}^{2}}[\boldsymbol{Y}^{T}\boldsymbol{Y}-\boldsymbol{Y}^{T}\boldsymbol{\Phi}^{T}\mathbf{A}^{-1}\boldsymbol{\Phi}\boldsymbol{Y}]\nonumber \\
 & -\frac{1}{2}\mathrm{log}|\mathbf{A}|+m\mathrm{log}\frac{m\sigma_{n}^{2}}{\sigma_{f}^{2}}-\frac{t}{2}\mathrm{log}2\pi\sigma_{n}^{2}.\label{eq:marginallikelihood}
\end{align}

\noindent The SSGP, therefore, uses $m$ optimal frequencies to approximate
the full GP, which reduces the computational complexity to $\mathcal{O}(tm^{2})$,
and provides computational efficiency if $m\ll t$. 
\begin{figure}[H]
\begin{raggedright}
\subfloat[full GP\label{fig:1-(a)}]{\begin{centering}
\includegraphics[scale=0.4]{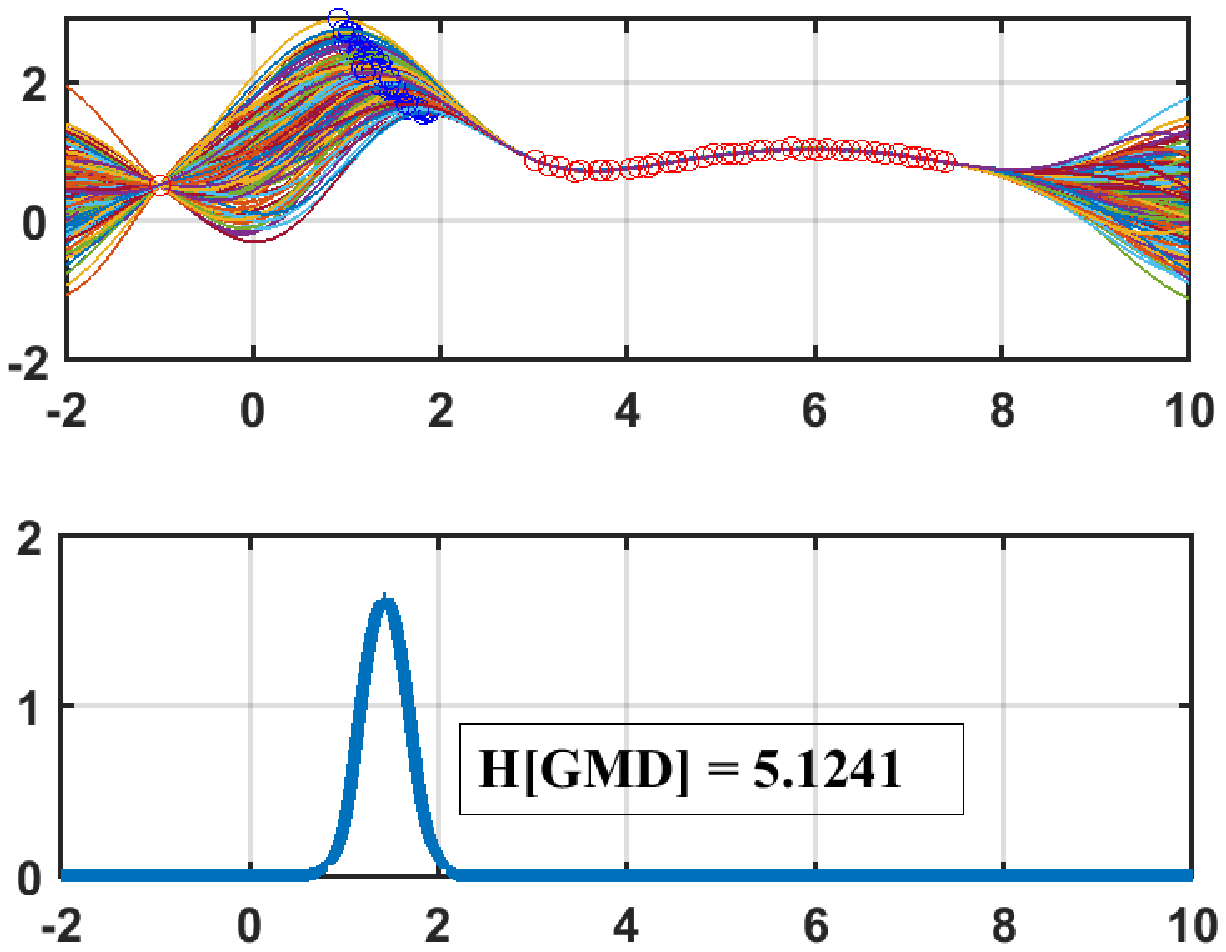}
\par\end{centering}
}\subfloat[SSGP\label{fig:1-(b)}]{\begin{centering}
\includegraphics[scale=0.4]{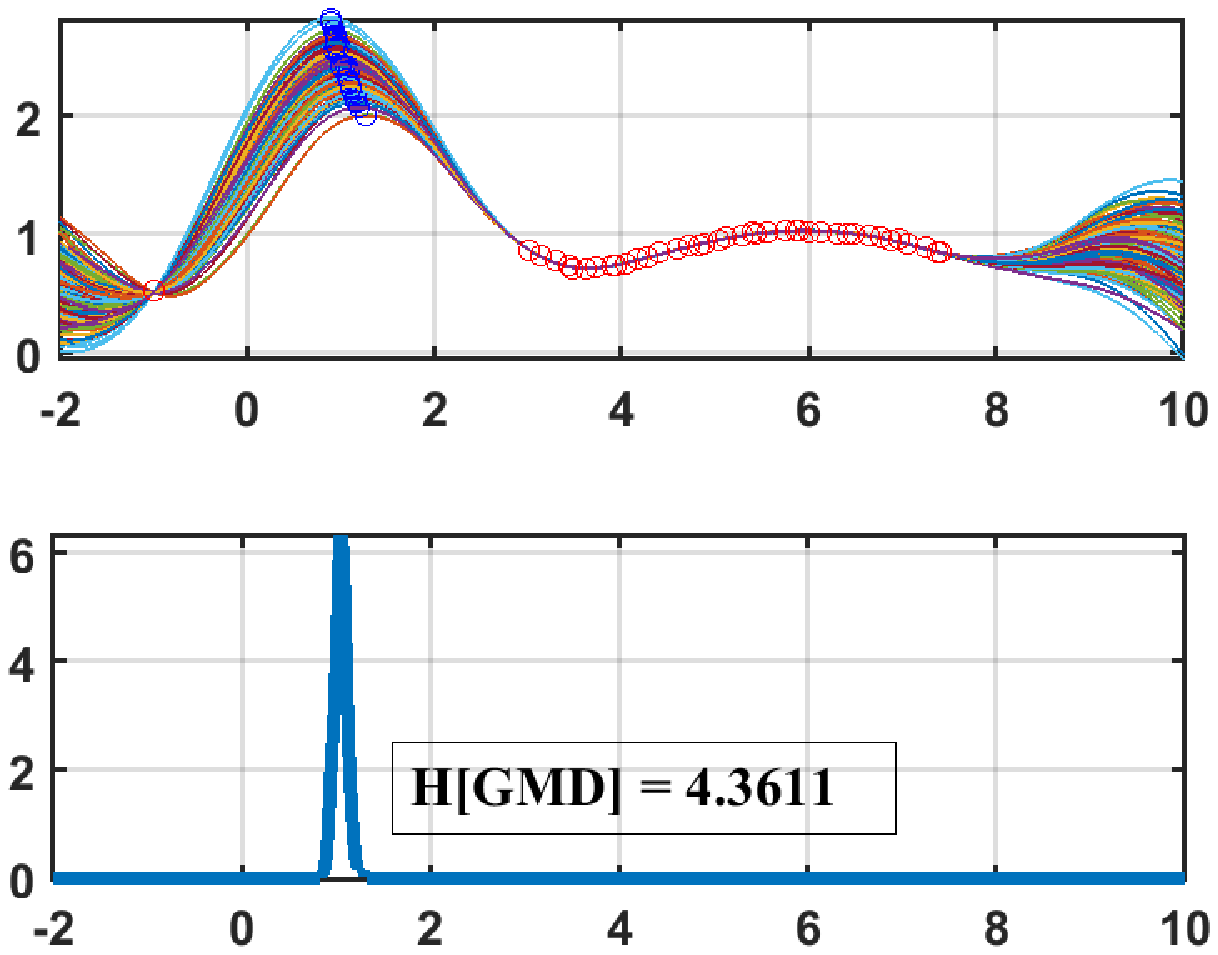}
\par\end{centering}
}
\par\end{raggedright}
\begin{raggedright}
\subfloat[RSSGP\label{fig:1-(c)}]{\centering{}\includegraphics[scale=0.41]{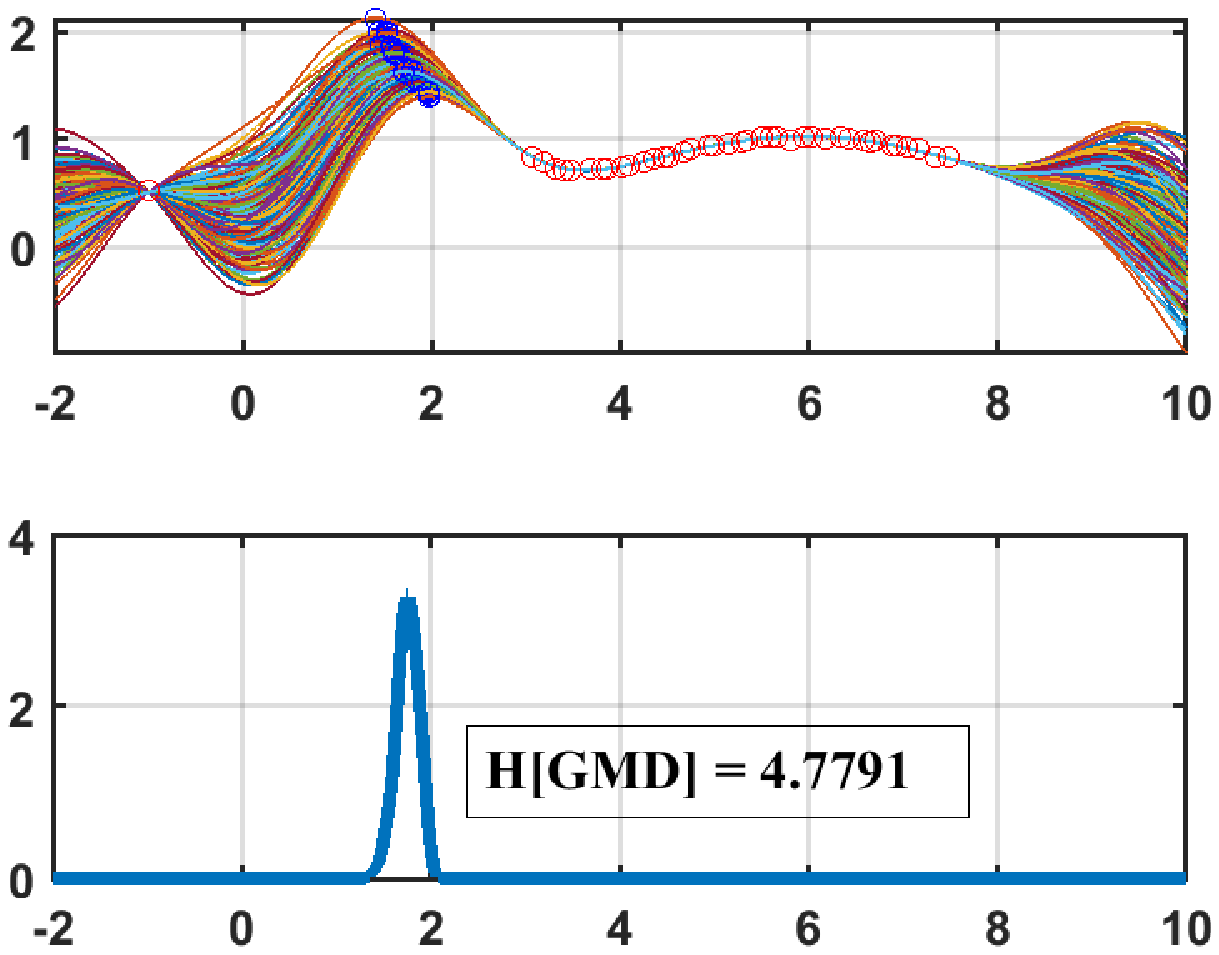}}\subfloat[RSSGP using EI as a proxy \label{fig:1-(d)}]{\begin{centering}
\includegraphics[scale=0.42]{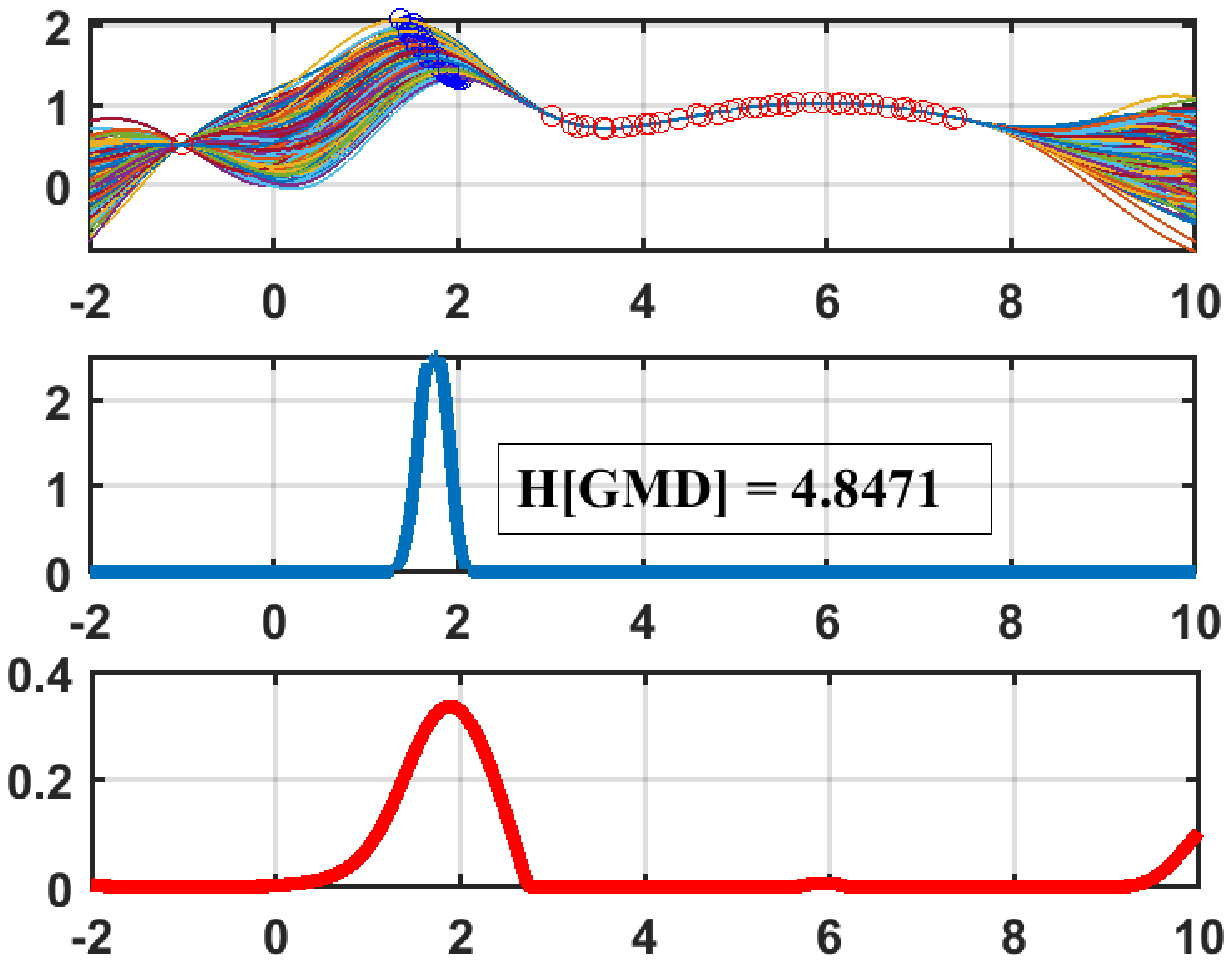}
\par\end{centering}
}
\par\end{raggedright}
\caption{\label{fig:1}(a)-(c) The visualization of overconfidence of SSGP
on the GMD. The \emph{upper} graphs show 200 posterior samples of
Sinc function, modeled by (a) full GP, (b) SSGP with 30 optimal frequencies,
and (c) RSSGP with 30 optimal frequencies. The red circle denotes
observation and the blue circle denotes the maximum location of a
posterior sample. The \emph{lower} graphs illustrate the resultant
GMD respectively. The $\mathrm{\mathbb{H}}\left[\mathrm{GMD}\right]$
is the entropy of the GMD. We can see the GMD of RSSGP is closer to
that of full GP than SSGP. (d) RSSGP with 30 optimal frequencies by
using the EI function as a proxy to the regularization. Its GMD is
at the middle and the EI function is at the bottom. }
\end{figure}

\section{Bayesian optimization using regularized sparse spectrum Gaussian
process }

The naive SSGP can be directly used for Bayesian optimization by replacing
the full GP. However, the SSGP was designed for a regression task,
which means that it assigns modeling resources equally over the whole
support space. Moreover, it leads to overconfidence on the GMD of
interest in BO. We illustrate the overconfidence of the SSGP in Figure
\ref{fig:1-(a)}-\ref{fig:1-(b)}, where the GMD of SSGP in \ref{fig:1-(b)}
(bottom) is narrower and sharper than that in the \ref{fig:1-(a)},
which is the GMD of the full GP. 

To overcome this overconfidence, we propose a novel and scalable sparse
spectrum Gaussian procesprobability of improvement,s model tailoring
for BO. Our approach involves maximizing a new loss function to select
the optimal spectrum frequencies. We design the loss function to include
the marginal likelihood in the SSGP and a regularization term, which
has the goal of minimizing the difference between the GMD of the full
GP and that of the proposed sparse approximation. We denote our proposed
sparse spectrum model as the regularized SSGP (RSSGP). For the sake
of convenience, we denote the GMD of the full GP as $p(\boldsymbol{x}^{*})$
and that of RSSGP as $q(\boldsymbol{x}^{*})$. 

We first discuss the choice for the regularization term. Whilst the
KL divergence $D_{KL}(q\,||\,p)$ seems to be the solution to measure
difference between two distributions, it is not feasible in our scenario
as we cannot access $p(\boldsymbol{x}^{*})$.  Nevertheless, the
property that the SSGP tends to be over-fitting implies that the entropy
for the GMD in SSGP would be smaller than that of the full GP.Therefore,
we can use the entropy of $q(\boldsymbol{x}^{*})$, or $\mathrm{\mathbb{H}}[q(\boldsymbol{x}^{*})]$
as the regularization term in the loss function that needs to be maximized.
In this way, the resultant sparse GP would minimize the difference
between $q(\boldsymbol{x}^{*})$ and $p(\boldsymbol{x}^{*})$. Formally,
the loss function in RSSGP is defined as 
\begin{equation}
\mathcal{L}=\mathrm{log}p(\boldsymbol{Y}|\varTheta)+\lambda\mathrm{log}\mathrm{\mathbb{H}}\left[q(\boldsymbol{x}^{*})\right],\label{eq:10}
\end{equation}
where the first term is the log marginal likelihood as Eq.(\ref{eq:marginallikelihood})
in SSGP, the second term is the entropy for the posterior distribution
of the global maximizer $q(\boldsymbol{x}^{*})$ and $\lambda$ is
the trade-off parameter. Now we can obtain $\varTheta$ by maximizing
the loss function
\begin{equation}
\varTheta=\mathrm{argmax}\mathrm{log}p(\boldsymbol{Y}|\varTheta)+\lambda\mathrm{log}\mathrm{\mathbb{H}}\left[q(\boldsymbol{x}^{*})\right].\label{eq:argmax}
\end{equation}
The questions break down to that how $q(\boldsymbol{x}^{*})$ can
be computed and how $q(\boldsymbol{x}^{*})$ is relevant to the spectrum
frequencies. Next, knowing there is no analytical form for $q(\boldsymbol{x}^{*})$,
we propose two methods to estimate $q(\boldsymbol{x}^{*})$. One is
Thompson sampling and the other is a sequential Monte Carlo approach
that takes less computation. We also propose a significantly computationally-efficient
approximation by treating the EI acquisition function as a proxy of
$q(\boldsymbol{x}^{*})$.

\subsection{Thompson sampling based approach}

In this section we demonstrate how to approximate $q(\boldsymbol{x}^{*})$
by following the work of \cite{40}. In Thompson sampling (TS), one
uses a linear model to approximate the function $f(\boldsymbol{x})=\phi(\boldsymbol{x})^{T}\bar{\boldsymbol{\theta}}$
where $\bar{\boldsymbol{\theta}}\backsim\mathcal{N}(\boldsymbol{0},\mathbf{I})$
is a standard Gaussian. Giving observed data $\mathcal{D}_{t}$, the
posterior of $\bar{\boldsymbol{\theta}}$ conditioning $\mathcal{D}_{t}$
is a normal $\mathcal{N}(\mathbf{A}^{-1}\boldsymbol{\Phi}^{T}\boldsymbol{Y},\mathbf{A}^{-1}\sigma_{n}^{2})$,
where $\mathbf{A}$ and $\boldsymbol{\Phi}$ have already been defined
in Eq.(\ref{eq:8}). Note that $\phi(\boldsymbol{x})$ is a set of
random Fourier features in the traditional TS while $\phi(\boldsymbol{x})$
is a set of $m$ pairs of symmetric Fourier features in our framework.

To estimate the global maximum distribution in RSSGP, we let $\phi_{i}$
and $\bar{\boldsymbol{\theta}}_{i}$ be a random set of $m$ pairs
of features and corresponding posterior weights. Both are sampled
according to the generative process above and they can be used to
construct a sampled function $f_{i}(\boldsymbol{x})=\phi_{i}(\boldsymbol{x})^{T}\bar{\boldsymbol{\theta}_{i}}$.
We can maximize this function to obtain a sample $\boldsymbol{x}_{i}^{*}$.
Once we have acquired sufficient samples, we use histogram based method
to obtain the probability mass function (PMF) over all $\boldsymbol{x}^{*}$,
denoted as $F(\boldsymbol{x}^{*})$. Then we estimate the entropy
via $\mathrm{\mathbb{H}}\left[q(\boldsymbol{x}^{*})\right]=-\sum_{i=1}^{L}F(\boldsymbol{x}_{i}^{*})\,\mathrm{log}F(\boldsymbol{x}_{i}^{*})$,
where $L$ is the number of samples. Since our RSSGP uses Fourier
features $\phi(\boldsymbol{x})$ to approximate a stationary kernel
function, and $q(\boldsymbol{x}^{*})$ also changes with applying
different Fourier features, therefore we can obtain the optimal features
by maximizing the combined term $\mathcal{L}$ in Eq.(\ref{eq:10}).
As a result, the selected optimal frequencies in RSSGP are not only
take care of posterior mean approximation, but also maximize the entropy
of $q(\boldsymbol{x}^{*})$. This is the key reason why we choose
SSGP as our base sparse method. Sparse models like FITC and VFE are
not capable with this idea since we cannot relate their sparse sets
to their GMDs due to the insufficient researches in this area.

We illustrate the GMD of RSSGP in Figure \ref{fig:1-(c)}. We can
see that it is closer to the GMD of the full GP than that of SSGP.
All the GMDs in Figure \ref{fig:1} are estimated via TS. 
\begin{figure}[H]
\centering{}\subfloat[MC performance \label{fig:2a}]{\centering{}\includegraphics[scale=0.4]{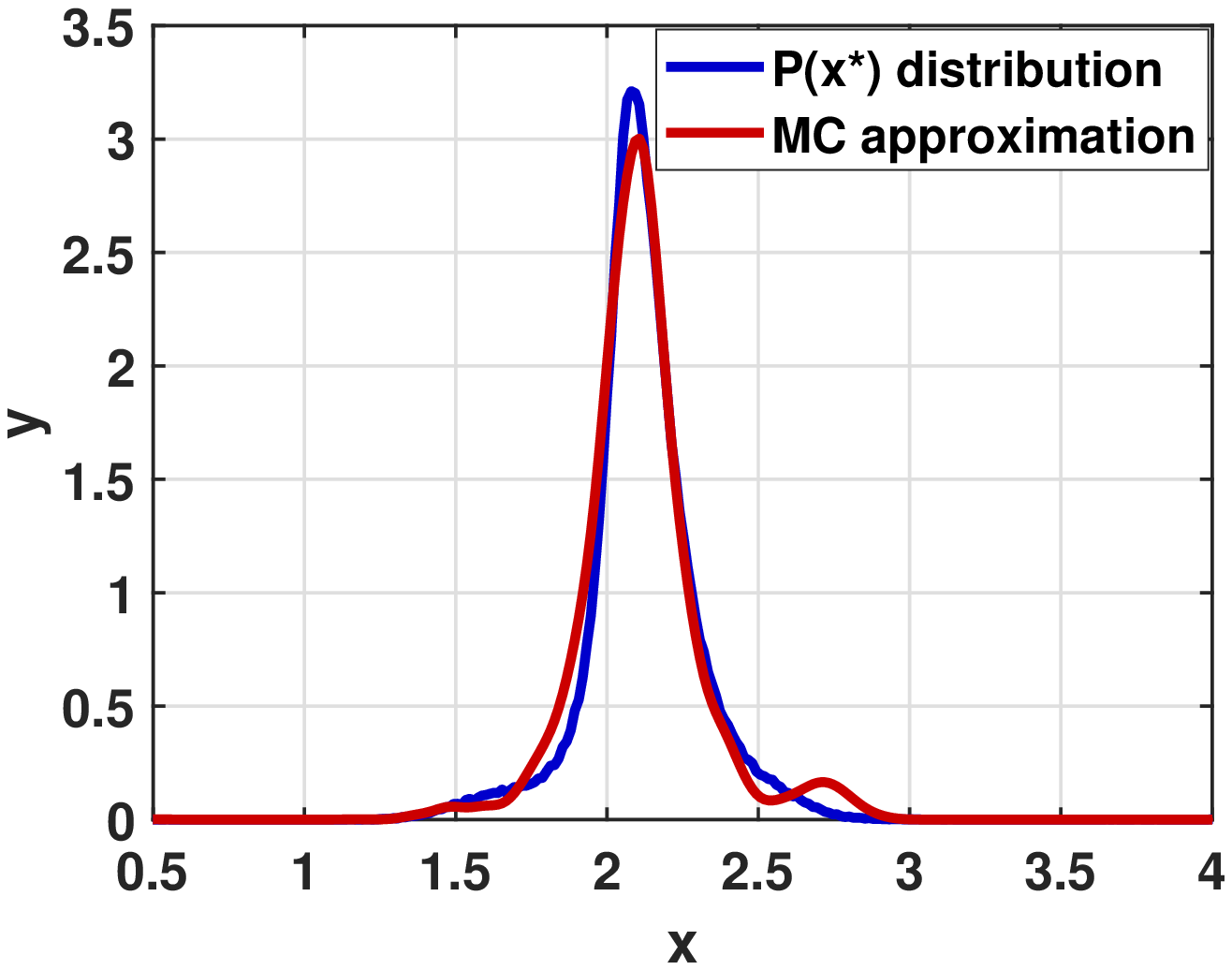}}\subfloat[TS performance \label{fig:2b} ]{\centering{}\includegraphics[scale=0.4]{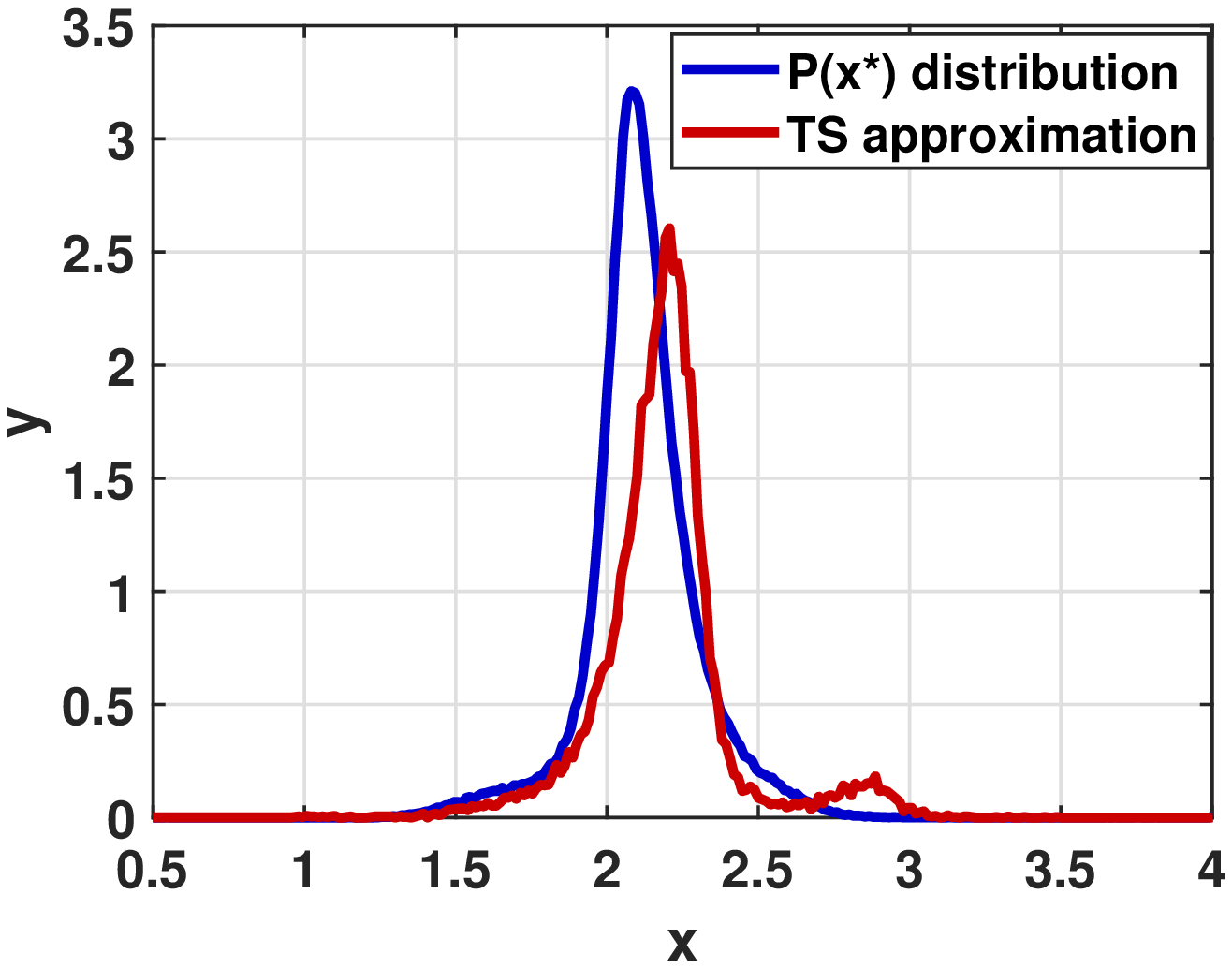}}\caption{MC approach vs TS approach. The blue lines show the reference $p(\boldsymbol{x}^{*})$
distribution, while the red lines illustrate the approximation using
MC (a) and TS (b) given the same running time. \label{fig:2}}
\end{figure}

\subsection{Monte Carlo based approach }

The estimation of $q(\boldsymbol{x}^{*})$ by TS often requires thousands
of samples (e.g, $L$), one of which involves the inversion of a $m\times m$
matrix. Inspired by a recent work \cite{bijl2016sequential} employing
sequential Monte Carlo algorithm to approximate the GMD, we develop
an intensive approach to estimate $q(\boldsymbol{x}^{*})$ in RSSGP
with significantly less computation. 

We start with $n_{p}$ particles at positions $\boldsymbol{\bar{x}}^{1},\ldots,\boldsymbol{\bar{x}}^{n_{p}}$.
Then we assign each particle a corresponding weight $\omega_{1},\ldots,\omega_{n_{p}}$.
Ultimately, these particles are supposed to converge to the maximum
distribution. At each iteration, we can approximate the $q(\boldsymbol{x}^{*})$
through kernel density estimation 
\begin{equation}
q(\boldsymbol{x}^{*}=\boldsymbol{x})\thickapprox\frac{\sum_{i=1}^{n_{p}}\omega_{i}k(\boldsymbol{x},\boldsymbol{\bar{x}}^{i})}{\sum_{i=1}^{n_{p}}\omega_{i}},\label{eq:13}
\end{equation}

\noindent where $k(\boldsymbol{x},\boldsymbol{\bar{x}}^{i})$ is the
approximated covariance function using $m$ features as in Eq.(\ref{eq:abc}). 

All the particles are sampled from the flat density distribution $v(\boldsymbol{x})=\beta$
at the beginning, so that they are randomly distributed across the
input space and the constant $\beta$ is nonzero. To obtain the maximum
position, we will challenge existing particles. We first sample a
number of $n_{c}$ challenger particles from a proposal distribution
$v'(\boldsymbol{x})$ and denote them as $\boldsymbol{\bar{x}}_{C_{1}},\ldots,\boldsymbol{\bar{x}}_{C_{n_{c}}}$.
To challenge an existing particle e.g. $\boldsymbol{\bar{x}}^{i}$,
we need to set up the joint distribution over $\boldsymbol{\bar{x}}^{i}$and
all challenger particles, which is a multivariate Gaussian distribution
\cite{bijl2016sequential}. We can subsequently generate a sample
$[\bar{f_{i}},\bar{f}_{C_{1}},\cdots,\bar{f}_{C_{n_{c}}}]^{T}$ from
the joint distribution. If the maximum value in the sample is greater
than $\bar{f_{i}}$, we replace $\boldsymbol{\bar{x}}^{i}$ with the
corresponding challenger particle. Otherwise, we retain the existing
particle. 

The challenger particle has an associated weight, which is often set
as the ratio of the initial distribution over the proposal distribution.
To speed up converge, we use the proposal distribution $v'(\boldsymbol{x})$
that is the mixture of the initial distribution and the current particle
distribution,
\begin{equation}
v'(\boldsymbol{x})=(1-\alpha)v(\boldsymbol{x})+\alpha q(\boldsymbol{x}^{*}=\boldsymbol{x})\label{eq:proposaldistribution}
\end{equation}

\noindent where $q(\boldsymbol{x}^{*}=\boldsymbol{x})$ is estimated
through Eq.(\ref{eq:13}) and $\alpha$ is trade-off parameter (e.g.,
$0.5$ in our experiments). To generate a challenger particle $\boldsymbol{\bar{x}}_{C_{1}}^{i}$,
we first select one of the existing particles e.g. $\boldsymbol{\bar{x}}^{i}$
according to the particle weights. Based on Eq.(\ref{eq:proposaldistribution}),
we then can sample $\boldsymbol{\bar{x}}_{C_{1}}^{i}$ from $k(\boldsymbol{x},\boldsymbol{\bar{x}}^{i})$
with the probability $\alpha$ or from the flat density distribution
$v(\boldsymbol{x})$ with the probability $1-\alpha$. Hence, the
challenger particle has a weight as
\begin{equation}
\omega_{C_{j}}^{i}=\frac{v(\bar{\boldsymbol{x}}_{C_{j}}^{i})}{\alpha k(\boldsymbol{\bar{x}}_{C_{j}}^{i},\boldsymbol{\bar{x}}^{k})+(1-\alpha)v(\bar{\boldsymbol{x}}_{C_{j}}^{i})}.\label{eq:15}
\end{equation}
Based on this information, we will challenge every particle once.
After each round, the systematic re-sampling \cite{kitagawa1996monte}
will be employed to make sure that all particles have the same weight
for the next round. This process stops till sufficient rounds. Thereafter,
we calculate the PMF of the particles and then estimate its entropy.

The Monte Carlo (MC) approach does not require a large matrix inversion
or nonlinear function optimization for the purpose of $q(\boldsymbol{x}^{*})$
approximation. Moreover, during the optimization process of optimal
features, $q(\boldsymbol{x}^{*})$ does not vary a lot with the change
of $\varTheta$. Therefore, most of the particles can be reused in
optimization process, significantly reducing computation cost. 

We demonstrate the superiority in Figure \ref{fig:2}. We denote the
GMD estimated from 50,000 TS samples of a full GP posterior on a 1$d$
function as our reference $p(\boldsymbol{x}^{*})$, showing as blue
lines in Figure \ref{fig:2a} and Figure \ref{fig:2b}. We give the
same running time ($0.5s$) to TS and MC approaches to approximate
the reference $p(\boldsymbol{x}^{*})$ respectively, showing as red
lines in Figure \ref{fig:2a} and Figure \ref{fig:2b}. We can see
that our MC approach successfully approximate the reference $p(\boldsymbol{x}^{*})$
while TS is not desirable. 

\subsection{Expected improvement acquisition function as a proxy}

To further reduce the computation, we propose to use EI function as
a proxy for $q(\boldsymbol{x}^{*})$. This choice is reasonable in
sense that they both measure the belief about the location of the
global maximum, and it can be seen from Figure \ref{fig:1-(d)} that
the GMD of full GP and the EI resembles closely. In this setting of
RSSGP, we can expect similar performance of capturing $q(\boldsymbol{x}^{*})$
information compare to the example showing in Figure \ref{fig:1-(c)}.
Since EI is a function, so we shall firstly use histogram based method
to acquire the PMF of EI and then to calculate the entropy. Although
it may inaccurate, in most of the cases we find it to work satisfactorily.

We use stochastic gradient descent to optimize Eq.(\ref{eq:argmax})
although alternatives are available. The proposed method is described
in Algorithm \ref{alg:BayesianOptimisation-2}.
\begin{algorithm}[H]
\caption{Regularised Sparse Spectrum Gaussian Process for Bayesian Optimization
\label{alg:BayesianOptimisation-2}}
1:$\mathbf{for}$ $n$ = $1$, $2$,...$t$ $\mathbf{do}$ 

2: $\,\,\,\,$Optimize Eq.(\ref{eq:argmax}) to obtain hyerpararameters
and optimal features

3: $\,\,\,\,$Fit the data $\mathcal{D}_{t}$ with RSSGP,

4: $\,\,$Suggest the next point $\boldsymbol{x}_{t+1}$ by maximising
$\boldsymbol{x}_{t+1}$ = $\text{argmax}\alpha_{EI}(\boldsymbol{x}|\mathcal{D})$,

5: $\,\,\,\,$Evaluate the function value $y_{t+1}$, 

6: $\,\,\,\,$Augment the observations$\mathcal{D}_{t}=\mathcal{D}_{t}\cup(\boldsymbol{x}_{t+1},y_{t+1})$. 

7: $\mathbf{end}$ $\mathbf{for}$ 
\end{algorithm}

\section{Experiments }

In this section, we evaluate our methods on optimizing benchmark functions,
an alloy design problem and hyperparameter tuning of machine learning
problems using transfer learning. We compare the following probabilistic
models used for Bayesian optimization:
\begin{itemize}
\item Full Gaussian process (\textbf{Full GP}) 
\item Sparse spectrum Gaussian process (\textbf{SSGP})
\item Our method 1: Regularized sparse spectrum Gaussian process with MC
estimation for $q(\boldsymbol{x}^{*})$ (\textbf{RSSGP-MC})
\item Our method 2: Regularized sparse spectrum Gaussian process with EI
approximation for $q(\boldsymbol{x}^{*})$ (\textbf{RSSGP-EI})
\item Variational Fourier features for Gaussian process using additive kernel
(\textbf{VFF-AK})
\item Variational Fourier features for Gaussian process using Kronecker
kernel (\textbf{VFF-KK})
\end{itemize}
In all settings, we use EI as the acquisition function in Bayesian
optimization and use the optimiser DIRECT \cite{61} to maximize the
EI function. We include both RSSGP-MC and RSSGP-EI in synthetic experiments.
We later only use RSSGP-EI due to its computational advantage and
the similar performance with RSSGP-MC. Given $d$-dimensional optimization
problems and $m$ frequencies, the size of inducing variables would
be $(2m)*d$ for VFF-AK and $(2m)^{d}$ for VFF-KK \cite{hensman2017variational}.
Thus, VFF-KK becomes almost prohibitively expensive for $d>2$ and
a large $m$.
\begin{figure}[H]
\begin{centering}
\subfloat[2d-Ackley\label{4-a}]{\begin{centering}
\includegraphics[scale=0.4]{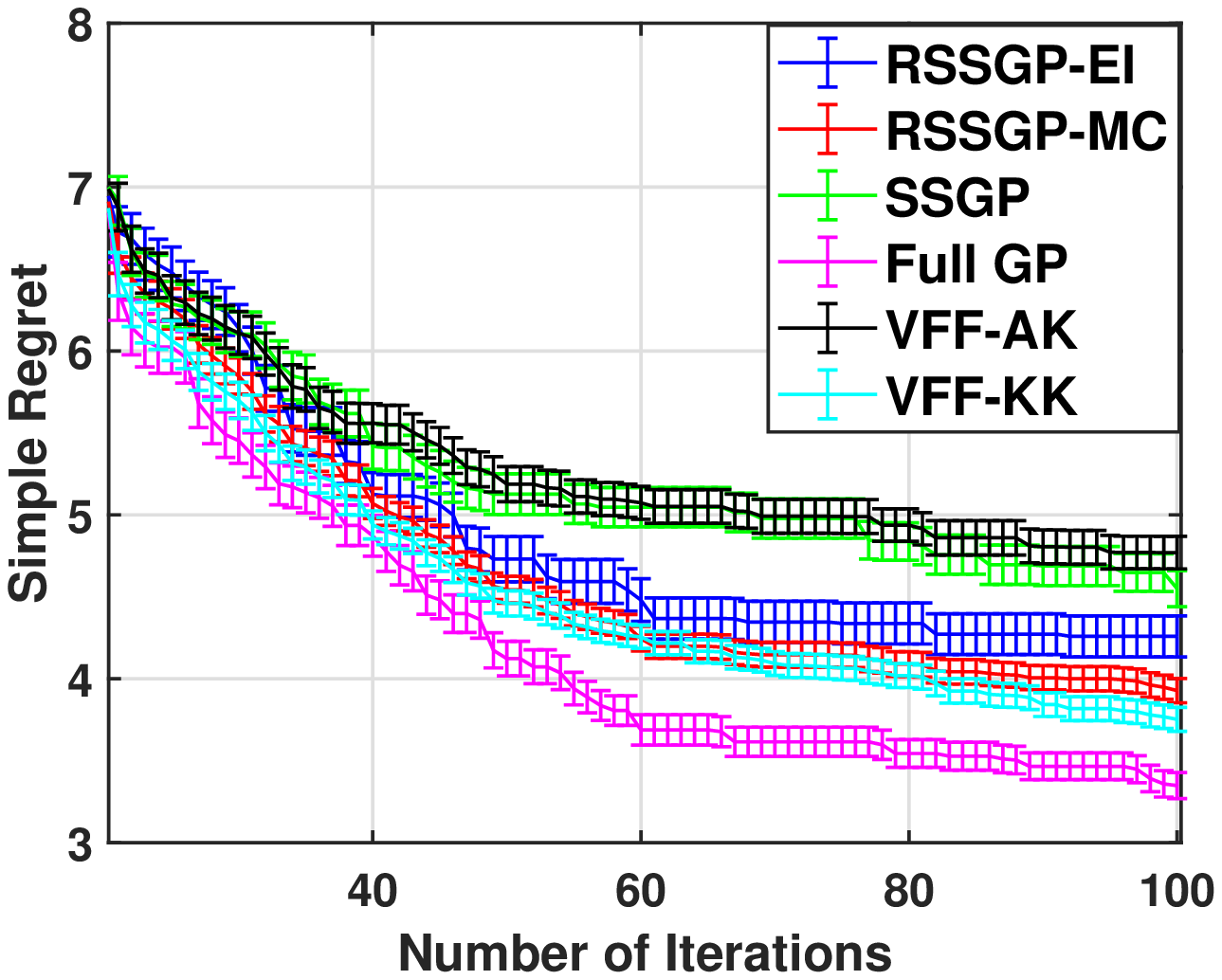}
\par\end{centering}
}\subfloat[6d-Hartmann\label{4-c}]{\begin{centering}
\includegraphics[scale=0.4]{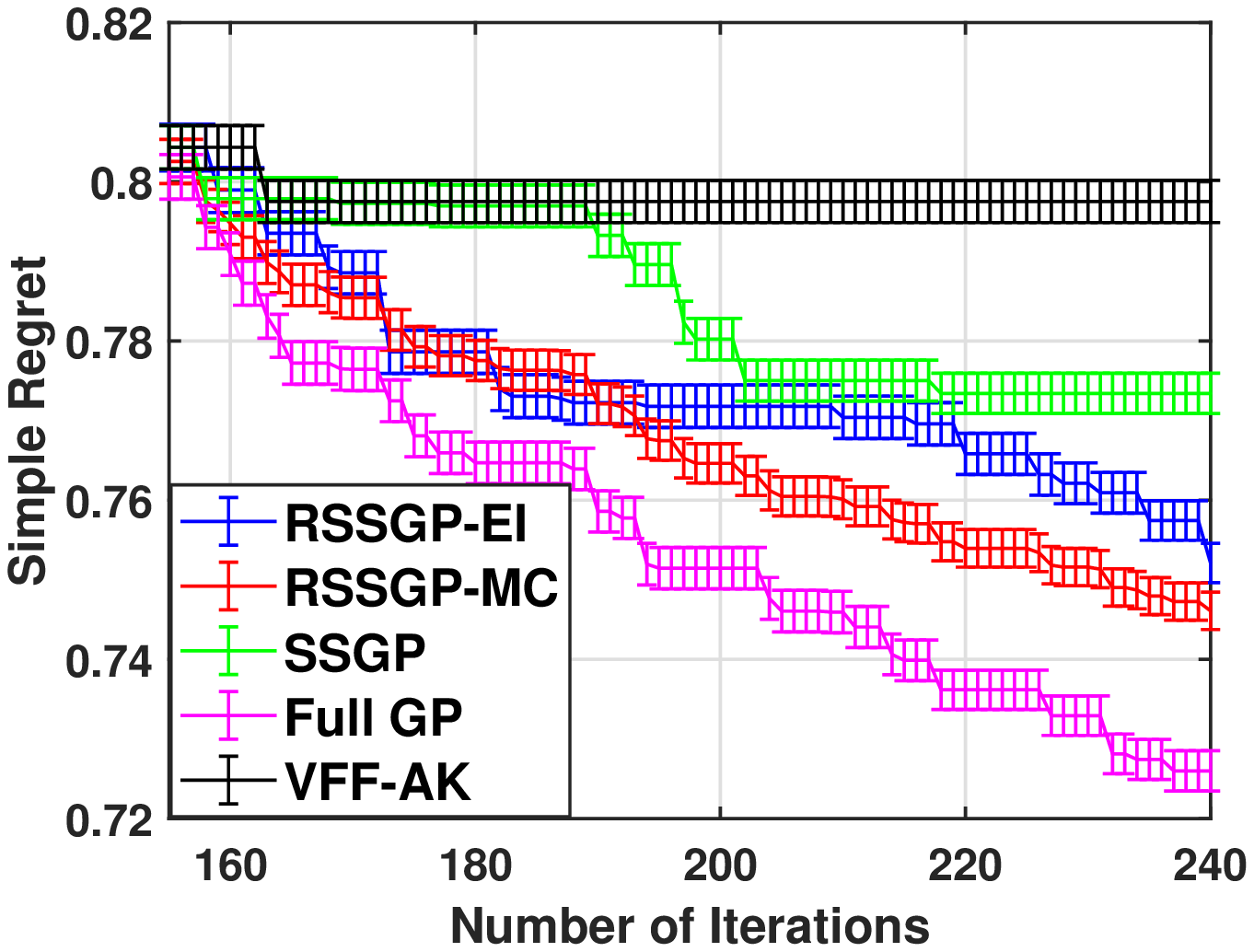}
\par\end{centering}
}
\par\end{centering}
\caption{Simple regret vs iterations for the optimization of (a) 2$d$ Ackley
and (b) 6$d$ Hartmann. The plots show the mean of minimum reached
and its standard error at each iteration. \label{fig:4}}
\end{figure}

\subsection{Optimizing benchmark functions}

We test on the following two benchmark functions: 
\begin{itemize}
\item 2$d$ Ackley function. Its global minimum is $f(\boldsymbol{x}^{\ast})=0$
and the search space is $[-10,10]^{2}$; 
\item 6$d$ Hartmann function. Its global minimum is $f(\boldsymbol{x}^{\ast})=-3.32237$
and the search space is $[0,1]^{6}$; 
\end{itemize}
We run each method for 50 trials with different initializations and
report the average simple regret along with standard errors. The simple
regret is defined as $r_{t}=f(\boldsymbol{x}^{*})-f(\boldsymbol{x}^{+})$,
where $f(\boldsymbol{x}^{*})$ is the global maximum and $f(\boldsymbol{x}^{+})=\max{}_{\boldsymbol{x}\in\{\boldsymbol{x}_{1:t}\}}f(\boldsymbol{x})$
is the best value till iteration $t$. In terms of kernel parameters,
we use the isotropic length scale, $\rho_{l}=0.5,\forall l$, signal
variance$\sigma_{f}^{2}=2$, and noise variance $\sigma_{n}^{2}=(0.01)^{2}$.
We empirically find that the proposed algorithms perform well when
the regularization term has the more or less scale with the log marginal
likelihood. Therefore, we set the trade-off parameter $\lambda=10$
for all of our methods. 

For the 2$d$ Ackley function, we start with 20 initial observations
and use 20 frequencies in all spectrum GP models. The experimental
result is shown in Figure \ref{4-a}. BO using FullGP performs best.
Both of our approaches (e.g., RSSGP-MC and RSSGP-EI) perform better
than SSGP. RSSGP-EI performs slightly worse than RSSGP-MC since it
only provides a rough approximation to the true global maximum distribution
but holds simplicity. VFF-KK performs well in a low dimensional problem
whilst VFF-AK performs worst. The use of additive kernel which does
not capture the correlation between dimensions may cause a bad performance
of VFF-AK.

For the 6$d$ Hartmann function, we start with 150 initial observations
and use 50 frequency features in all spectrum GP models. Similar results
as the 2$d$ Ackley function can be seen in Figure \ref{4-c}. We
did not run VFF-KK on this 6$d$ function optimization due to a huge
size of inducing variables mentioned before.
\begin{figure}[H]
\begin{centering}
\includegraphics[scale=0.5]{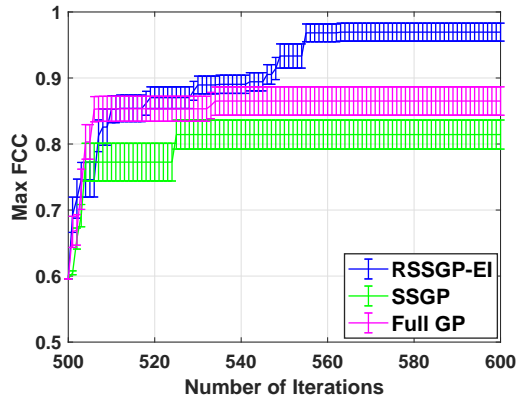}
\par\end{centering}
\caption{Alloy optimization-FCC at 15-dimensions. \label{fig:3d}}
\end{figure}

\subsection{Alloy optimization }

In the joint project with our metallurgist collaborators, we aim to
design an alloy with a micro-structure that contains as much fraction
of FCC phase as possible. We use a thermodynamic simulator called
ThermoCalc \cite{andersson2002thermo}. Given a composition of an
alloy, the simulator can compute thermodynamic equilibrium and predict
the micro-structure of the resultant alloy using CALPHAD \cite{saunders1998calphad}
methodology. In this experiment, the search space is a 15 dimensional
combination of the elements: Fe, Ni, Cr, Ti, Co, Al, Mn, Cu, Si, Nb,
Mo, W, Ta, C, N. For each composition, ThermoCalc provides the amount
of FCC in terms of volume fraction. The best value of volume fraction
is 1. Since ThermoCalc takes around 10 minutes per composition to
compute volume fraction, it fits perfectly in our notion of semi-expensive
functions. We use 500 initial points and 50 sparse features and run
5 different trials with different initial points. The results in Figure
\ref{fig:3d} shows BO with RSSGP-EI performs the best over all three
methods. We found that the covariance matrix of the full GP easily
became ill-conditioned in the presence of a large number of observations,
and hence, fails to be inverted properly, being ended up harming the
BO.
\begin{figure}[H]
\begin{centering}
\subfloat[LiverDisorders as Target\label{fig:6a}]{\centering{}\includegraphics[scale=0.4]{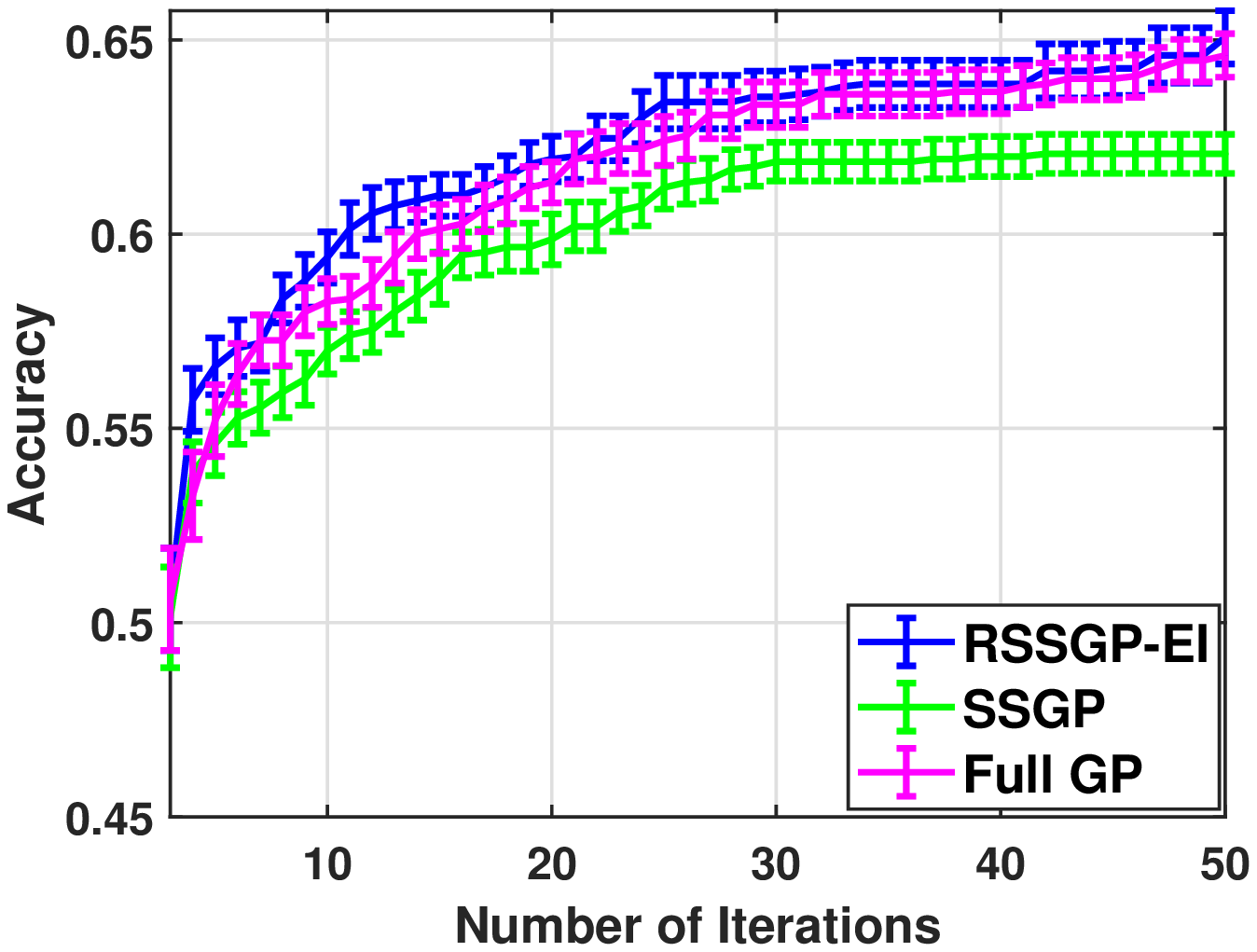}}\subfloat[Madelon as Target\label{fig:6b}]{\centering{}\includegraphics[scale=0.4]{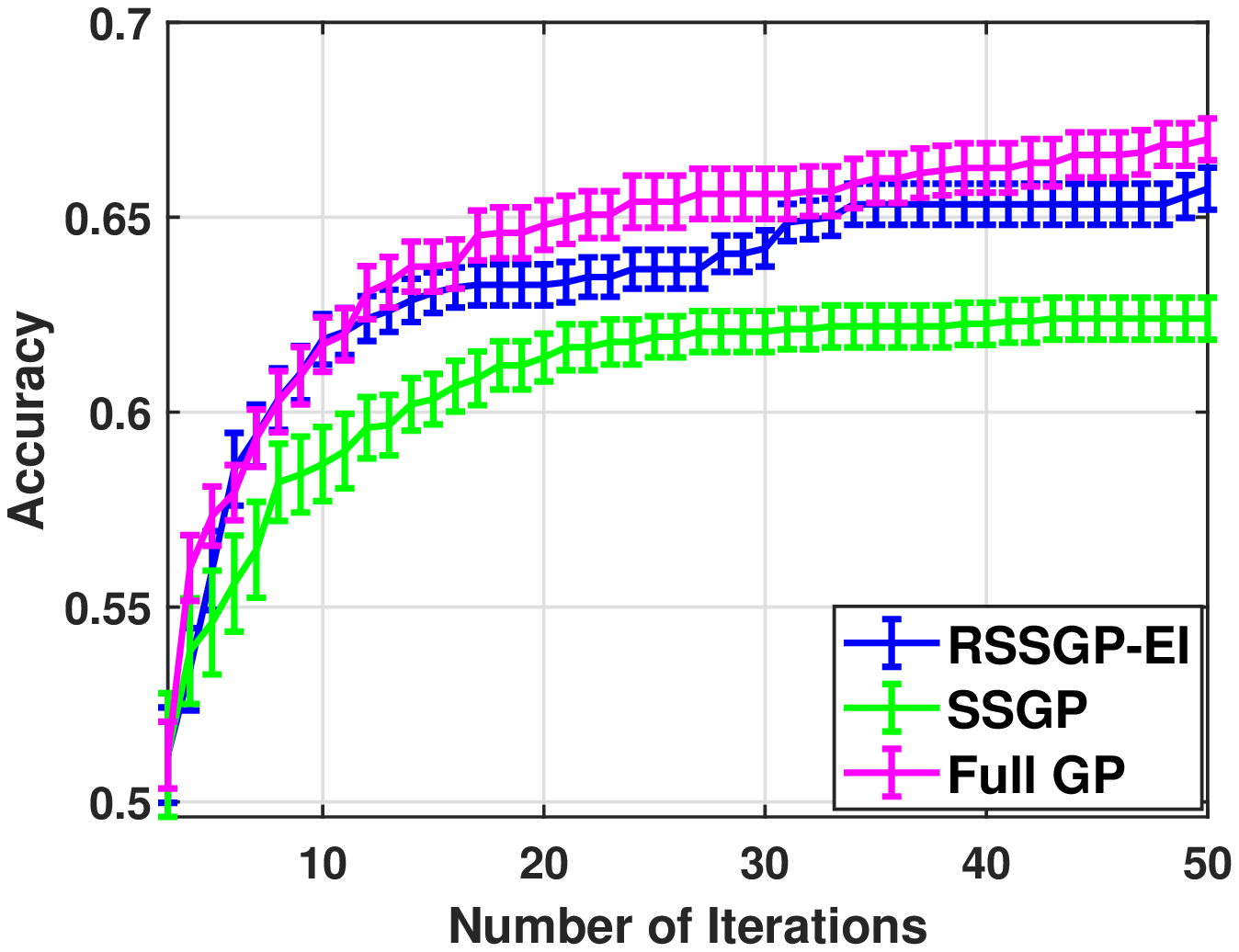}}
\par\end{centering}
\begin{centering}
\subfloat[Mushroom as Target\label{fig:6c}]{\centering{}\includegraphics[scale=0.4]{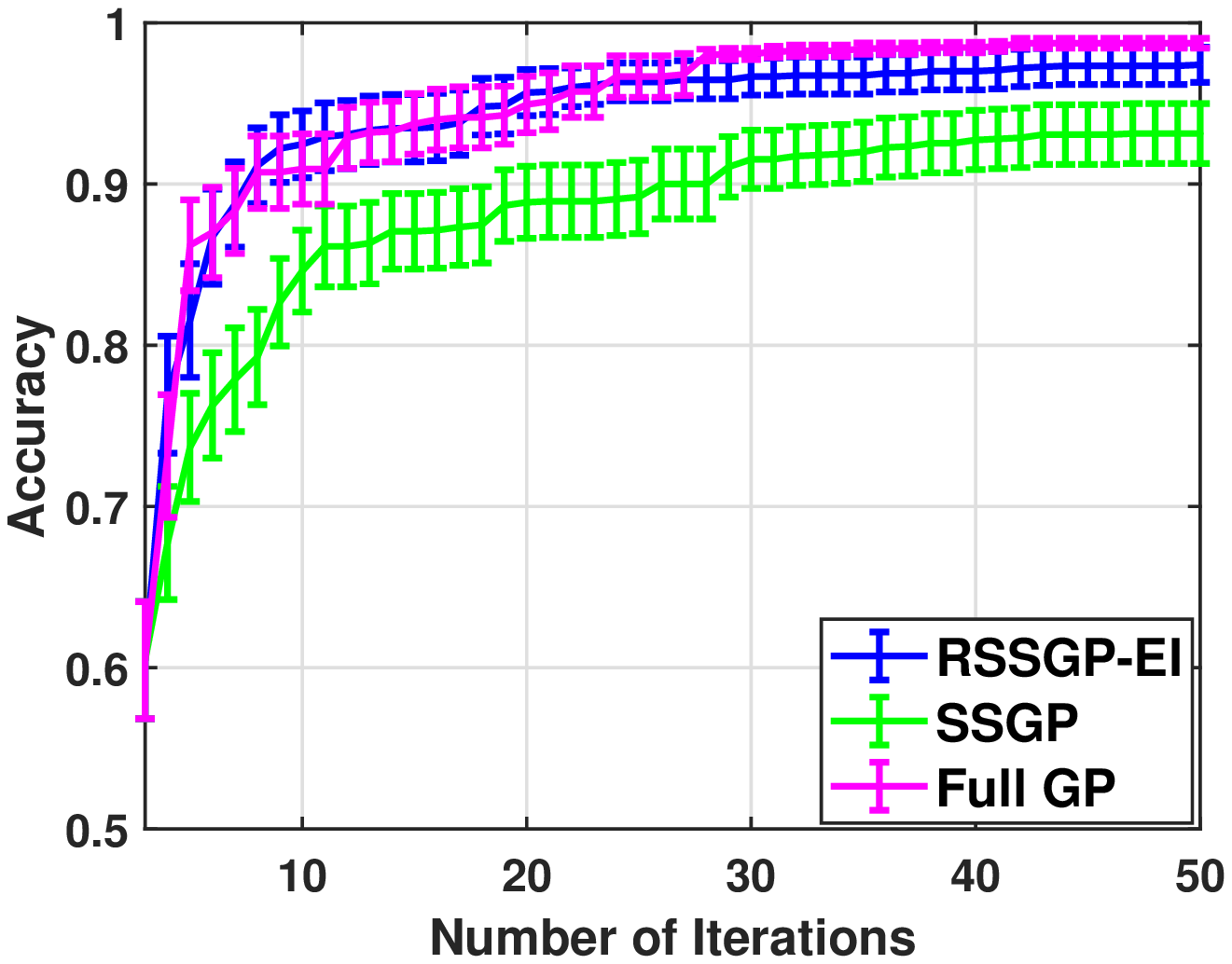}}\subfloat[BreastCancer as Target\label{fig:6d}]{\centering{}\includegraphics[scale=0.4]{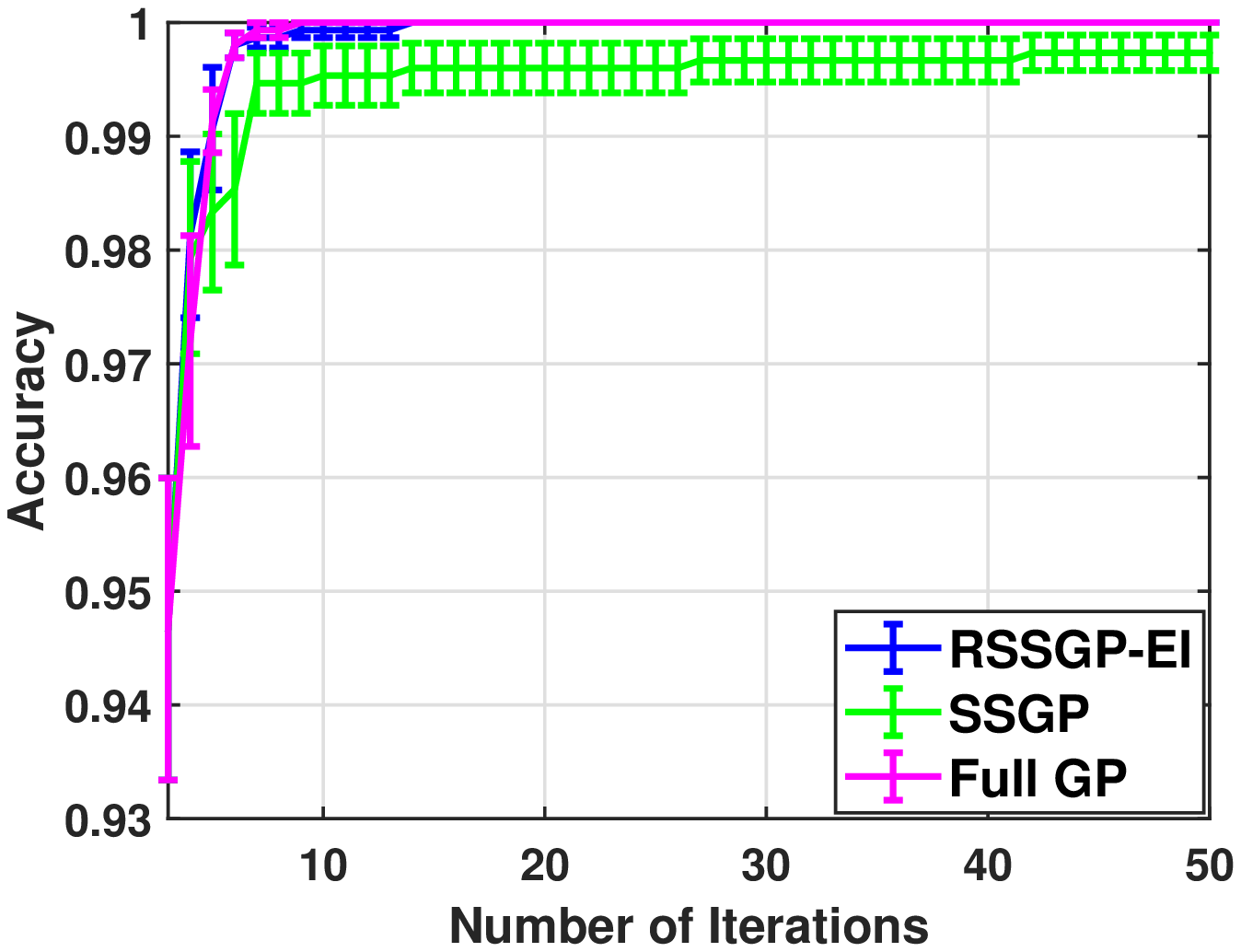}}
\par\end{centering}
\caption{Hyperparameters tuning for SVM classifier by transfer learning. The
plots show the maximal accuracy reached till the current iteration
by RSSGP-EI (blue), SSGP (green) and by FullGP (mauve). Error bar
indicates the standard error.\label{fig:6} }
\end{figure}

\subsection{Transfer learning on hyperparameter tuning }

Transfer learning in the context of Bayesian optimization pools together
observations from the sources and the target to build a covariance
matrix in the GP. In this case when the number of sources is large
or/and the number of existing observations per source is large, the
resultant covariance matrix can be quite huge, demanding a sparse
approximation. We conduct experiments for tuning hyperparameters of
Support vector machine (SVM) classifier in a transfer learning setting.
We use the datasets: LiverDisorders, Madelon, Mushroom and BreastCancer
from UCI repository \cite{Dua:2017} and construct four transfer learning
scenarios. For each scenario, we use 3 out of 4 datasets as the source
tasks, and the rest one as the target task. We randomly generate 900
samples of hyperparameters and the corresponding accuracy values from
each source task. We also randomly generate 3 initial samples from
the target task. As a result, we have 2703 initial observations to
build the combined covariance matrix. Following the framework \cite{joy2019flexible}
where source points are considered as noisy observations for the target
function, we add a higher noise variance (3 times of that in target
observations) to 2700 source observations . This allows us to use
the same covariance function to capture the similarity between the
observations from both source and target tasks. We optimize two hyperparameters
in SVM which are the cost parameter ($C$) and the width of the RBF
kernel ($\gamma$). The search bounds for the two hyperparameters
are $C$ =$10^{\lambda}$ where $\lambda\in[-3,3]$, and $\gamma=10^{\omega}$
with $\omega\in[-3,0]$, respectively, and we optimize $\lambda$
and $\omega$. We run each scenario 30 trials with different initializations.
The results are showed in Figure \ref{fig:6}. We can see that in
all scenarios BO with RSSGP-EI outperforms the naive SSGP. We note
that the covariance matrix of full GP for transfer learning tasks
does not always suffer from ill-conditioning since the source observations
have a higher noise. Therefore, we can see BO with the full GP works
well from the results.

\section{Conclusion}

In this paper we proposed a new regularized sparse spectrum Gaussian
process method to make it more suitable for Bayesian optimization
applications. The original formulation results in an over-confident
GP. BO using such an over-confident GP may fare poorly as the correct
uncertainty prediction is crucial for the success of BO. We propose
a modification to the original marginal likelihood based estimation
by adding the entropy of the global maximum distribution induced by
the posterior GP as a regularizer. By maximizing the entropy of that
distribution along with the marginal likelihood, we aim to obtain
a sparse approximation which is more aligned with the goal of BO.
We show that an efficient formulation can be obtained by using a sequential
Monte Carlo approach to approximate the global maximum distribution.
We also experimented with the expected improvement acquisition function
as a proxy to the global maximum distribution. Experiments on benchmark
functions and two real world problems show superiority of our approach
over the vanilla SSGP method at all times and even better than the
usual full GP based approach at certain scenarios.

\end{document}